\documentclass[10pt,twocolumn,letterpaper]{article}

\usepackage[pagenumbers]{cvpr} 
\usepackage{booktabs, multirow, threeparttable}
\usepackage[table]{xcolor}
\usepackage{tcolorbox}
\usepackage{setspace}

\definecolor{cvprblue}{rgb}{0.21,0.49,0.74}
\usepackage[pagebackref,breaklinks,colorlinks,allcolors=cvprblue]{hyperref}

\def\paperID{11167} 
\def\confName{CVPR}
\def\confYear{2026}

\title{Advancing Multimodal Judge Models through a Capability-Oriented Benchmark and MCTS-Driven Data Generation}

\author{Zeyu Chen \quad Huanjin Yao \quad Ziwang Zhao \quad Min Yang\textsuperscript{*} \\
{\tt\small * Correspondence to yangminbupt@outlook.com}
}

\begin{document}
\maketitle
\section*{\centering Abstract}
\label{sec:abstract}
\textit{Using Multimodal Large Language Models (MLLMs) as judges to achieve precise and consistent evaluations has gradually become an emerging paradigm across various domains. Evaluating the capability and reliability of MLLM-as-a-judge systems is therefore essential for ensuring trustworthy assessment. Existing judge benchmarks categorize samples by task types but fail to capture the fundamental judgment capabilities required for reliable evaluation. In this work, we introduce M-JudgeBench, a ten-dimensional capability-oriented benchmark designed to comprehensively assess the judgment abilities of MLLMs. Our benchmark decomposes evaluation into pairwise Chain-of-Thought (CoT) comparison, length bias avoidance, and process error detection tasks, jointly covering ten fine-grained subtasks. This design enables diagnosis of model reliability across reasoning styles, response lengths, and cross-model variations. Systematic evaluation uncovers the systematic weaknesses in existing MLLM-as-a-judge systems. To address this issue, we further propose Judge-MCTS, a data construction framework generating pairwise reasoning trajectories with various correctness and length. Using Judge-MCTS, we construct an MCTS-augmented dataset and train M-Judger, a series of strong judge models. Extensive experiments demonstrate the superiority of M-Judger on existing judge benchmarks as well as M-JudgeBench. Overall, our work establishes a more principled foundation for evaluating MLLM-as-a-judge through M-JudgeBench and Judge-MCTS framework, paving the way for future research on judge model evaluation and capability-driven judge training.}
\section{Introduction}
\label{sec:intro}

Multimodal large language models (MLLMs) have recently achieved remarkable progress across diverse perception and reasoning tasks\cite{yin2024survey}. As these models become increasingly capable, the challenge has shifted from producing multimodal outputs to evaluating them\cite{zhang2023gpt}. In this context, judge model plays a pivotal role in assessing the quality of MLLM responses and guiding alignment training\cite{li-etal-2025-generation, chen2024mllmasajudge}. A powerful judge model that can accurately rank model responses is capable of generating high-quality preference training data. It improves the efficacy of post-training methods, such as direct preference optimization\cite{NEURIPS2023_a85b405e, xiong2025llava}.

Existing judge benchmarks for evaluating MLLM-as-a-judge, such as VL-RewardBench\cite{li_vl-rewardbench_2025_upd}, Multimodal RewardBench\cite{yasunaga_multimodal_2025}, and JudgeAnything\cite{pu_judge_2025_upd}, primarily organize evaluation data by task types (e.g., image understanding, image generation, mathematical reasoning, and general knowledge task). Although this approach provides task-level coverage, it fails to measure the core judgmental abilities that define whether a model truly behaves like a reliable evaluator. From a human perspective, an effective judge should possess several essential capabilities:
(1) Accurately distinguish quality differences among responses with the same answering style, ensuring consistent preference selection even under highly similar reasoning formats. (2) Generalize across diverse response styles from different models or individuals, reliably assessing answer quality regardless of linguistic patterns, verbosity, or reasoning habits. (3) Maintain fairness when comparing Chain-of-Thought\cite{wei2022chain} (CoT) responses of different lengths, from concise ShortCoT output to detailed LongCoT reasoning with explicit thinking processes. (4) Avoid being misled by logically coherent yet factually incorrect reasoning chains, prioritizing correctness while not relying solely on surface-level reasoning structure. (5) Identify fine-grained reasoning issues, including visual misinterpretations, logical fallacies, and incidental mistakes such as copy or transcription errors. These core abilities reflect the human perspective of judgment, yet they are largely overlooked by existing judge benchmarks and training schemes, leaving a critical gap between task-type coverage and capability-oriented evaluation.

\begin{figure}[htbp]
  \centering
  \includegraphics[width=0.8\linewidth]{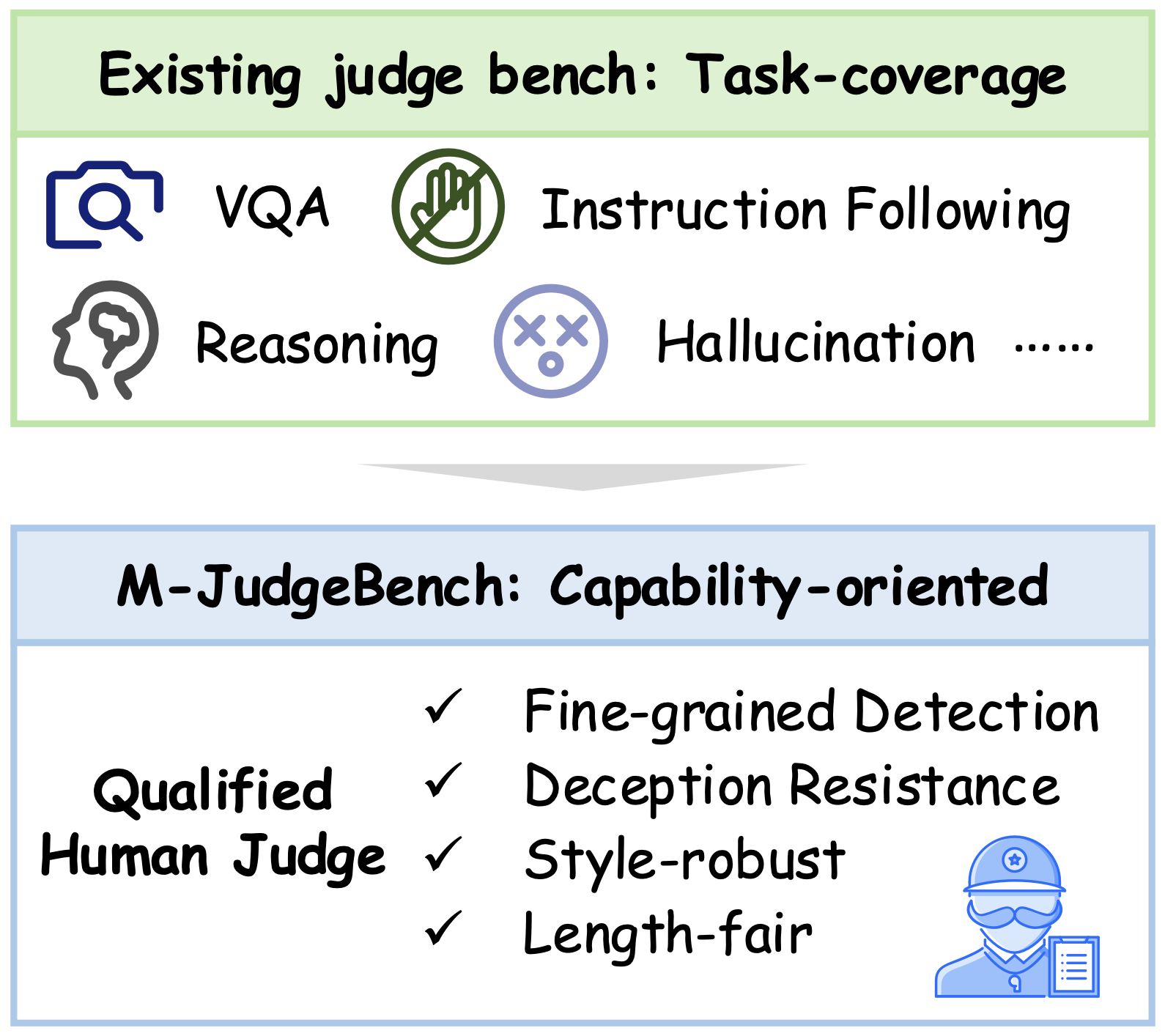}
  \caption{Comparison between existing judge benchmarks and M-JudgeBench. M-JudgeBench is designed with an emphasis on evaluating judgment capabilities.}
  \label{fig:comparison}
\end{figure}

To address this issue, we introduce M-JudgeBench, a capability-oriented MLLM judge benchmark inspired by how humans assess answer quality (Figure \ref{fig:comparison}). We decompose judgment into two essential dimensions: result error judgment, which determines correctness across responses with different reasoning styles or lengths, and process error detection, which inspects the quality of the reasoning chain even when the final answer is correct. By disentangling these complementary cognitive factors, our design enables a more fine-grained and principled analysis of MLLM judge behavior, revealing failure modes that are not captured by task-type categorization alone.

Comprehensive evaluation reveals that current small-scale MLLMs, including general-purpose ones like Qwen3-VL\cite{qwen3technicalreport, bai_qwen25-vl_2025_upd} and InternVL-3.5\cite{wang2025internvl3_5}, as well as specialized judge models such as InternLM-XComposer2.5-Reward\cite{zang-etal-2025-internlm}, Unified Reward\cite{wang_unified_2025}, UnifiedReward-Think\cite{wang_unified_2025-1_upd}, and R1-Reward\cite{zhang_r1-reward_2025}, still suffer from systematic deficiencies. Specifically, their training data are typically synthesized by diversifying question categories but not by modeling the underlying cognitive abilities of judgment. As a result, these models often exhibit poor sensitivity to reasoning errors, limited adaptation to diverse response styles, and non-trivial length bias. Notably, these persistent limitations are not confined to smaller models, but are also observed in more capable proprietary models such as the GPT series.

To enhance judge model capability with minimal additional data cost, we introduce Judge-MCTS, an MCTS-based data construction framework. Monte Carlo Tree Search (MCTS) is a powerful framework for efficient exploration and decision making\cite{MCTS_ori}. Starting from reasoning seed data, step-level rollouts are performed to generate structured reasoning trajectories, where each sampled node represents a valid intermediate state in the reasoning process\cite{xie2024monte, yao2025mulberry}. This procedure naturally produces responses spanning four categories, including long-correct, long-error, short-correct, and short-error, enabling the creation of fine-grained pairwise supervision signals. Diverse and contrastive reasoning pairs allow models to better distinguish subtle differences across reasoning lengths and styles, while effectively reducing the inherent length bias present in current judge models.

Our main contributions are summarized as follows:

\textbf{Capability-Oriented Benchmark.} We propose M-JudgeBench that systematically evaluates MLLM-as-a-judge systems, revealing the systematic weaknesses of existing small-scale judge models. Beyond providing a high-quality benchmark, M-JudgeBench offers a general and scalable approach to strengthen existing judge benchmarks, elevating their difficulty and evaluative power.

\textbf{MCTS-Based Data Construction Method.} We introduce the Judge-MCTS framework to generate step-wise, correctness-labeled reasoning trajectories, enabling pairwise preference training for enhanced judge models. 

\textbf{A Series of Strong Judge Models.} Leveraging Judge-MCTS, we augment multiple base models and develop the M-Judger series. The inclusion of MCTS-augmented data leads to consistent performance improvements on existing judge benchmarks as well as M-JudgeBench.

\section{M-JudgeBench Construction}
\label{sec:benchmark}
\subsection{Capability-Oriented Evaluation Framework}
\begin{figure*}[htbp]
  \centering
  \includegraphics[width=0.9\textwidth]{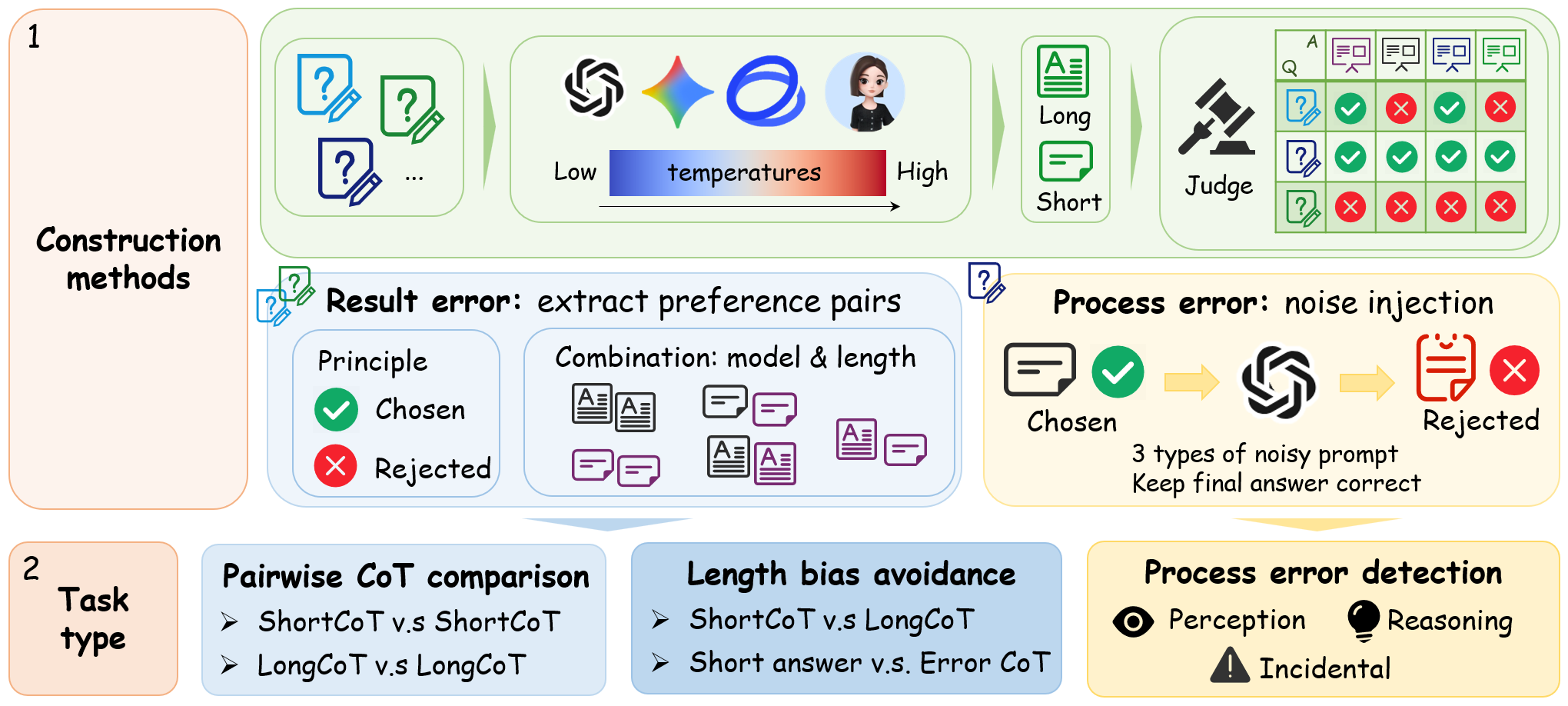}
  \caption{Overview of M-JudgeBench. The figure illustrates the data construction methods and resulting task types in M-JudgeBench. Result-error pairs are derived from rollouts of different models with varied temperatures and reasoning lengths, while process-error data are produced by controlled noise injection preserving correct answers. These yield three main task types and ten subtasks in total.}
  \label{fig:overview}
\end{figure*}

While existing judge benchmarks primarily categorize samples by task types, M-JudgeBench introduces a capability-oriented evaluation inspired by how humans assess answer quality. As illustrated in Figure \ref{fig:overview}, we design two essential dimensions and three main tasks, including result error and process error. 

Result error judgment measures the ability to distinguish correct and incorrect responses across varying answer lengths and styles. It encompasses several types of pairwise comparisons: ShortCoT pairs within the same model or across models, LongCoT pairs within the same model or across models, and length-bias probing pairs. Together, these settings enable a systematic examination of both the fidelity of reasoning and the length bias of judge models. Process error detection focuses on reasoning robustness when the final answer is correct. It includes three representative error types: (1) visual understanding mistakes, (2) logical reasoning fallacies, and (3) incidental process errors (e.g., spelling or transcription). Benchmark data examples are provided in Supplementary Material. 

\subsection{Data Generation Steps}
We collect high-quality open-source benchmark data as the seed sources for constructing M-JudgeBench, including MMMU\cite{yue_mmmu_2024_upd}, MMMU-Pro\cite{yue_mmmu-pro_2025_upd}, MMStar\cite{chen_are_2024_upd}, MMReason\cite{yao_mmreason_2025_upd}, M3CoT\cite{chen_m3cot_2024_upd}, MathVision\cite{wang_measuring_2024_upd}, and MathVerse\cite{zhang_mathverse_2024_upd} (Supplementary Material for data statistics). 

Then, we generate diverse responses by performing rollouts across multiple models and decoding temperatures (0.01, 0.5, 1.0, 1.5, 2.0). The response sources include Gemini 2.5 Pro\cite{comanici2025gemini} (ShortCoT), GPT-4.1 (ShortCoT), Seed1.6-VL\cite{guo_seed15-vl_2025_upd} (ShortCoT and LongCoT), and GLM-4.5V\cite{team_glm-45v_2025} (ShortCoT and LongCoT), providing variations in both reasoning style and length. Next, we use GPT-4.1 to extract the final answers from the CoT outputs and evaluate their correctness against ground-truth labels.

Following this multi-step curation pipeline, each question is associated with multiple candidate responses of different correctness levels, source models, and reasoning lengths, enabling the construction of our three capability-oriented benchmark tasks.



\subsubsection{Pairwise CoT Comparison}
To evaluate the capability of judging correctness under reasoning-style variations, we select cases where both a correct and an incorrect response are available from the rollouts. The correct one is labeled as the chosen and the incorrect one as the rejected.

\noindent\textbf{CoT rollouts from different model.} This setting examines whether the judge model can generalize across heterogeneous reasoning styles. For each question, multiple responses are sampled using different temperatures from various strong MLLMs, resulting in correct–incorrect response pairs. Positive and negative responses in a pair originate from different backbone models, often exhibiting noticeable stylistic differences such as phrasing preference, detail level, or structural organization. Successfully selecting the correct response here reflects robustness to style variance across models and authors.

\noindent\textbf{CoT rollouts from the same model.} In contrast, this setting focuses on fine-grained discrimination within a consistent reasoning style. Both responses are generated by the same model with different sampling temperatures, making their surface form and logic organization highly similar. Correctly identifying the better response requires detecting subtle reasoning flaws rather than relying on stylistic cues, thus posing a challenging judgment scenario.

For both categories above, we construct two length-based pair types: LongCoT v.s. LongCoT and ShortCoT v.s. ShortCoT, allowing simultaneous evaluation of model adaptability to reasoning length and stylistic variation. Consequently, this task comprises four subtasks in total.
\subsubsection{Length Bias Avoidance}
The length bias avoidance task examines whether a judge model can remain neutral when comparing responses of significantly different lengths. To comprehensively evaluate this bias across practical settings, we include two complementary task forms.

\noindent\textbf{ShortCoT v.s. LongCoT.} This setting parallels the Pairwise CoT Comparison task but explicitly imposes a substantial length gap between the positive and negative samples. Depending on the rollout results, we include two subcases:
(1) correct LongCoT as chosen v.s. incorrect ShortCoT as rejected, and
(2) correct ShortCoT as chosen v.s. incorrect LongCoT as rejected.
This evaluates whether a judge can reliably prioritize factual correctness over verbosity or vice versa, without being misled by answer length.

\noindent\textbf{Correct short answer v.s. incorrect CoT.}
We further include hard cases where all rollout models fail to provide a correct reasoning chain. For such instances, we compose a concise correct answer, typically in the form of a short direct solution such as "Final Answer: X" as the chosen response. The rejected response is a longer but incorrect CoT generated by a strong model. This setup enforces the judge model to favor factual correctness even in the extreme case where the correct answer lacks reasoning details, while the incorrect one is structurally rich and seemingly persuasive.

\subsubsection{Process Error Detection}
The process error detection task evaluates whether a judge model can detect flawed or low-quality reasoning even when the final answer is correct. This dimension targets the deeper cognitive skill of assessing reasoning validity rather than relying solely on outcome correctness. We categorize process error issues into three subtypes, enabling judgment beyond surface-level correctness.

\noindent\textbf{Visual Perception Errors.} Misinterpretation of visual elements or spatial relationships, leading to incorrect intermediate conclusions despite deriving the correct final answer.

\noindent\textbf{Logical Reasoning Fallacies.} Invalid deductive steps, such as contradiction, circular dependency, or unsupported inference, hidden within a complete reasoning chain.

\noindent\textbf{Incidental Mistakes.} Minor but undesirable issues (e.g., spelling slips, symbol transcription errors, or unit inconsistencies) that do not change the final answer but reflect low-quality execution.

We construct this benchmark by selecting relatively easy visual reasoning questions where multiple models consistently produce correct final answers. For each such instance, we inject only one specific type of process error into a response using a carefully designed noisy prompt\cite{wang2024self, whitehouse2025j1}, while ensuring that the final answer remains correct (Supplementary Material for prompt templates). The clean reasoning chain is labeled as the chosen response, and the perturbed one as rejected. This setup forces the judge model to scrutinize and validate the entire reasoning process rather than relying solely on final-answer correctness.

\subsection{Statistics of M-JudgeBench}
To ensure high-quality contrastive samples, we further apply strict exact-match filtering: only pairs in which the extracted preferred answer matches the correct label and the rejected answer differs from it are preserved. 

\begin{figure}[!htbp]
  \centering
  \includegraphics[width=1.0\linewidth]{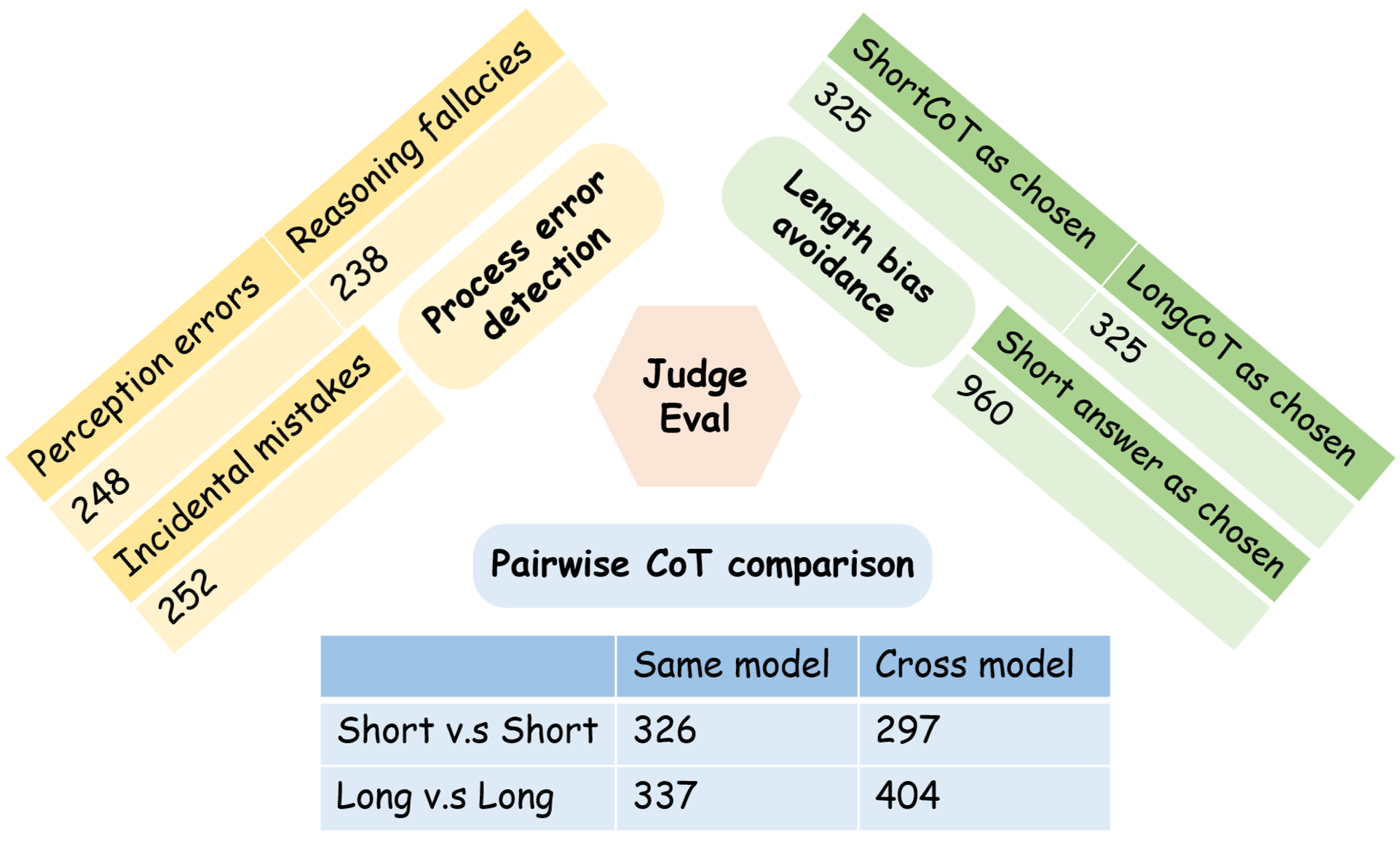}
  \caption{Data statistics of M-JudgeBench. }
  \label{fig:statistics}
\end{figure}

Finally, M-JudgeBench consists of 3,712 carefully curated multimodal instances in total. It covers three main categories and ten subtasks, including pairwise CoT comparison (1,364 pairs), length bias avoidance (1,610 pairs), and process error detection (738 pairs). We provide an overview of the data composition of M-JudgeBench, as illustrated in Figure \ref{fig:statistics}. It provides comprehensive and balanced coverage of core judgment capabilities that existing MLLM judge benchmarks overlook.

\section{M-Judger Training}
\label{sec:model}

\subsection{Open-Source Training Data Collection}
We collect pairwise training data from a diverse set of multimodal and text-only sources, covering task types including reasoning, instruction following, hallucination detection, visual question answering, mathematics, and coding. Specifically, our open-source preference mixture integrates high-quality datasets including MMPR\cite{wang_enhancing_2025}, MMIF\cite{ding_mm-ifengine_2025_upd}, RLAIF-V\cite{yu_rlaif-v_2024_upd}, POVID\cite{zhou_aligning_2024}, MIA-DPO\cite{liu_mia-dpo_2025_upd}, PDS-DPO\cite{wijaya_multimodal_2024}, UnifiedReward-img\cite{wang_unified_2025}, Skywork-Reward\cite{liu_skywork-reward_2024_upd}, StepDPO\cite{lai_step-dpo_2024}, and Ling-Coder-DPO\cite{codefuse_every_2025}. We modify the prompt design of Multimodal RewardBench and Unified Reward, and convert model comparison outputs into a pairwise ranking format. Specifically, we set "Answer X is better" as the target response for MLLMs to generate.

During model training and ablation experiments, we first apply strict data filtering to avoid information leakage. For instance, questions overlapping with evaluation benchmarks (e.g., POVID and RLAIF-V samples included in VL-RewardBench) are removed to avoid information leakage and ensure fair evaluation. Then, we downsample the original mixed dataset to reduce the overall data volume while preserving category diversity, resulting in approximately 142k pairwise training samples. Statistics of training data are provided in Supplementary Material. 
\subsection{Judge-MCTS Data Construction Framework}
\label{sec:mcts_data_construction}
Training a robust judge model requires diverse and fine-grained supervision signals. However, existing pairwise data generation pipelines focus on increasing the volume of data by task type. We apply the Monte Carlo Tree Search (MCTS) method to synthesize specialized and diverse reasoning trajectories, enriching the training data with process-level diversity. This method explores a wide tree of possible reasoning paths for a given question, allowing for the systematic generation of preference pairs that vary in reasoning length and complexity. 

We employ the MCTS algorithm based on the OmegaPRM\cite{luo_improve_2024} framework to generate a pairwise preference dataset. The high quality seed data are obtained from ThinkLite-VL\cite{wang_sota_2025_upd} and MM-K12\cite{du_mm-prm_2025}.

\subsubsection{Reasoning Rollout via MCTS}
Given a multimodal reasoning problem consisting of an image \(I\) and a question \(q\), we represent the reasoning process as a tree where each node corresponds to an intermediate reasoning step \(s_t\). Starting from the initial state, MCTS iteratively explores the space of reasoning actions, such as "identify object", "infer relation", or "apply formula". At each step, the node is expanded based on a policy guided by the base model's output probability and a value function estimating the correctness of the reasoning path.

Each rollout produces a complete reasoning path \(\pi = \{s_1, s_2, \dots, s_T\}\), where \(T\) is the number of steps in the reasoning process. We retain both high-value and low-value trajectories, thus obtaining four distinct types of final responses. The following four types are used to construct a wide range of reasoning pairs for model training.
\begin{itemize}
    \item \textbf{Short-Correct (SC)}: Brief but correct reasoning.
    \item \textbf{Short-Error (SE)}: Short reasoning with mistakes.
    \item \textbf{Long-Correct (LC)}: Long and correct reasoning path.
    \item \textbf{Long-Error (LE)}: Verbose reasoning with subtle or explicit errors.
\end{itemize}

\subsubsection{Pairwise Data Construction}
MCTS-based rollouts generate four structured response types: short-correct (SC), short-error (SE), long-correct (LC), and long-error (LE). These responses enable a systematic construction of chosen–rejected pairs:
\[
\text{Pair} = \{(\text{SC}, \text{SE}), (\text{LC}, \text{LE}), (\text{SC}, \text{LE}), (\text{LC}, \text{SE})\}.
\]

By pairing responses differ in both reasoning quality and length, the judge model is trained not only to verify correctness but also to evaluate the soundness and efficiency of reasoning paths. This helps the model distinguish superficially plausible yet flawed long reasoning from concise and valid solutions, ultimately improving its ability to detect subtle errors that are not reflected in the final answer alone.

\subsection{M-Judger Training Methodology}

\subsubsection{Stage1. Supervised Fine-Tuning (SFT).} We construct the Stage-1 judge models by supervised fine-tuning multiple initial backbones, using a mixture of diverse open-source pairwise preference datasets. To assess the contribution of MCTS data, we additionally train enhanced SFT variants for each backbone by injecting a proportion of MCTS-augmented pairwise reasoning samples into the open-source mixture, resulting in M-Judger-SFT.

\subsubsection{Stage2. Reinforcement Learning (RL).} Starting from the non-enhanced Stage-1 SFT models for each backbone, we further fine-tune the model using DAPO\cite{yu_dapo_2025}, resulting in M-Judger-RL. The RL dataset is constructed by mixing MCTS-augmented pairwise reasoning samples and high-quality open-source preference samples in equal proportion. This approach allows us to evaluate whether introducing a small amount of MCTS data at the RL stage can effectively enhance the judge models across different base architectures.

\section{Experiments}
\label{sec:experiments}

\subsection{Experimental Setup}
\subsubsection{Model Evaluation}
\noindent\textbf{Evaluated models}
To comprehensively assess cross-model generalization and judgment robustness, we evaluate a diverse set of models, which can be categorized into three groups: (1) closed-source general-purpose MLLMs, including Gemini 2.5 Pro, GPT-5 (in both thinking and no-thinking modes), GPT-4.1, GPT-4.1-mini, GPT-4o\cite{hurst2024gpt}, and Seed1.6-VL; (2) open-source general-purpose MLLMs, including GLM-4.5V (in both thinking and no-thinking modes), Qwen2.5-VL-7B-Instruct\cite{bai_qwen25-vl_2025_upd}, MiMo-VL-7B-SFT-2508\cite{team_mimo-vl_2025}, LLaVA-v1.6-Mistral-7B\cite{liu_improved_2024_upd}, InternVL3.5 (4B/8B), and Qwen3-VL-Instruct (2B/4B/8B); (3) open-source specialized judge models, including Unified Reward, UnifiedReward-Think, and R1-Reward.

\noindent\textbf{Evaluation metrics.} All results are reported using pairwise accuracy, computed as the percentage of cases in which the model correctly selects the preferred response in a binary comparison setup.

\noindent\textbf{Prompt settings.} We adopt two evaluation protocols aligned with prior judge models. For open-source judge models (Unified Reward, UnifiedReward-Think, and R1-Reward), we follow the original prompt settings described in their respective papers, with an additional instruction appended: \textit{"Please prioritize selecting the response with the most accurate answer as chosen, then analyze whether the reasoning process is thorough and correct"}. For other models (general-purpose MLLMs, models enhanced by Judge-MCTS), we design the prompt template according to the task types defined in M-JudgeBench and require the model to directly output its judgment without exposing intermediate reasoning steps (detailed in Supplementary Material). 

\subsubsection{Model Training}
\noindent\textbf{MCTS-data Generation.} We adopt the MM-PRM framework to construct reasoning rollouts. Qwen2.5-VL-7B-Instruct serves as the base rollout model, while Seed1.6-VL is used as the judge model to guide rollout selection. For each prompt, four rollouts are generated with a maximum search count of 50 and a temperature setting of 1.0.

\noindent\textbf{Supervised Fine-Tuning.} We conduct full-parameter SFT using Qwen3-VL-Instruct (4B/8B) as base models on the constructed pairwise preference dataset, as they achieve SOTA performance among models of comparable size on M-JudgeBench (Section \ref{sec:main_results}). SFT is conducted under two data settings: one using only 142k open-source samples (statistics in Supplementary Material), and another combining these samples with 13k MCTS-augmented samples. The models are trained with a batch size of 128, a learning rate of 2.5e-6, and for one epoch. Additionally, to comprehensively assess the generality of Judge-MCTS, four base models (Qwen3-VL-2B-Instruct, LLaVA-v1.6-Mistral-7B, Unified Reward, and InternVL3.5-8B) are also fine-tuned under the same settings for ablation studies. 

\noindent\textbf{Reinforcement Learning.}
For RL, we use the SFT models trained solely on open-source data (without MCTS augmentation) as the base models. We further fine-tune these models using the DAPO algorithm with a hybrid reward combining format and result reward. The RL dataset consists of 13k MCTS-augmented data and 16k open-source preference data (sampled from opensource pairwise mixture) mixed in equal proportion. During optimization, we set the actor clipping ratio between 0.20 and 0.28, and use a learning rate of 1e-6. The model is trained for three epochs.
\subsection{Main Results}\label{sec:main_results}
\subsubsection{Evaluation of MLLMs on M-JudgeBench}
Table \ref{tab:eval_results_all} summarizes the performance of representative models on our proposed benchmark. Our key findings are summarized as follows.

\begin{table*}[!htbp]
\centering
\caption{Performance evaluation of existing models on M-JudgeBench. Short/Long-same/cross: ShortCoT or LongCoT pair from the same model or different models, using correct as chosen and error as rejected; Ans as ch.: correct short answer v.s. incorrect CoT; Short as ch.: correct ShortCoT v.s. incorrect LongCoT; Long as ch.: correct LongCoT v.s. incorrect ShortCoT; Perception: Visual Perception Errors; Reasoning: Logical Reasoning Fallacies; Incidental: Incidental Mistakes.}
\setlength{\tabcolsep}{2pt}
\label{tab:eval_results_all}
\resizebox{\linewidth}{!}{
\begin{tabular}{lccccccccccc}
\toprule
\multirow{2}{*}{\textbf{Model}} & \multicolumn{4}{c}{\textbf{Pairwise CoT comparison}} & \multicolumn{3}{c}{\textbf{Length bias avoidance}} & \multicolumn{3}{c}{\textbf{Process error detection}} & \multirow{2}{*}{\textbf{Overall}} \\
\cmidrule(lr{0.1em}){2-5}\cmidrule(lr{0.1em}){6-8}\cmidrule(lr{0.1em}){9-11}
& \textbf{Short-same} & \textbf{Short-cross} & \textbf{Long-same} & \textbf{Long-cross} & \textbf{Ans as ch.} &  \textbf{Short as ch.} & \textbf{Long as ch.} & \textbf{Perception} & \textbf{Reasoning} & \textbf{Incidental} &  \\
\midrule
\multicolumn{12}{l}{\textit{Closed-source general-purpose MLLMs}} \\
Gemini2.5pro (2025-06-05) & \textbf{73.62} & \textbf{87.21} & \textbf{70.03} & \textbf{72.52} & 23.33 & 61.23 & 74.46 & \textbf{99.19} & 97.48 & 92.46 & \textbf{64.76} \\
GPT-5-chat (2025-08-07) & 61.04 & 72.73 & 56.68 & 61.14 & 9.06 & 42.15 & 76.92 & 95.56 & 97.90 & 80.16 & 53.85 \\
GPT-5-think (2025-08-07) & 56.75 & 67.68 & 58.46 & 62.62 & 20.63 & 47.38 & 72.00 & 96.77 & 97.48 & 82.14 & 56.60 \\
GPT-4.1 (2025-04-14) & 56.13 & 60.94 & 54.60 & 59.16 & 4.38 & 33.54 & \textbf{79.38} & 93.55 & 97.06 & 83.73 & 50.38 \\
GPT-4.1-mini (2025-04-14) & 57.36 & 58.59 & 53.12 & 60.64 & 8.75 & 32.31 & 71.08 & 89.92 & 91.60 & 81.75 & 49.89 \\
GPT-4o (2024-08-06) & 54.91 & 63.64 & 59.05 & 57.43 & 4.38 & 60.62 & 52.62 & 90.32 & 97.06 & 70.24 & 49.60 \\
Seed1.6-VL (2025-06-15) & 60.43 & 70.71 & 60.53 & 61.63 & 9.17 & 48.92 & 68.62 & 97.58 & 98.74 & 98.02 & 55.33 \\
\midrule
\multicolumn{12}{l}{\textit{Open-source general-purpose MLLMs}} \\
GLM-4.5V (106B-A12B) & 57.06 & 69.02 & 51.04 & 49.26 & 6.77 & 46.15 & 64.00 & 93.15 & 96.64 & 68.25 & 48.98 \\
GLM-4.5V-think (106B-A12B) & 61.35 & 77.78 & 51.34 & 52.48 & 7.92 & 34.15 & 70.15 & 97.58 & \textbf{99.58} & \textbf{99.21} & 52.80 \\
LLaVA-v1.6-Mistral-7B & 46.31 & 47.81 & 48.51 & 46.90 & \textbf{35.52} & 52.47 & 45.54 & 53.63 & 57.14 & 48.41 & 45.66 \\
Qwen2.5-VL-7B-Instruct & 46.01 & 52.53 & 44.81 & 46.78 & 4.38 & 46.46 & 53.85 & 54.84 & 58.40 & 41.27 & 39.63 \\
MiMo-VL-7B-SFT-2508 & 52.45 & 68.69 & 41.54 & 33.17 & 2.08 & 37.23 & 58.46 & 81.05 & 88.65 & 73.81 & 41.69 \\
InternVL3.5-4B & 57.67 & 66.67 & 50.15 & 54.70 & 3.54 & 61.85 & 40.92 & 77.42 & 80.25 & 53.57 & 44.77 \\
InternVL3.5-8B & 55.83 & 60.27 & 57.86 & 52.48 & 5.73 & 52.62 & 58.77 & 85.48 & 87.82 & 59.52 & 47.31 \\
Qwen3-VL-2B-Instruct & 46.93 & 59.93 & 60.53 & 53.22 & 8.02 & 59.08 & 45.85 & 79.44 & 87.82 & 52.78 & 45.99\\
Qwen3-VL-4B-Instruct & 58.59 & 71.72 & 55.79 & 59.41 & 3.54 & \textbf{76.92} & 40.31 & 89.11 & 92.44 & 73.81 & 50.00 \\
Qwen3-VL-8B-Instruct & 57.66 & 80.81 & 58.75 & 61.88 & 2.71 & 65.23 & 52.00 & 90.32 & 95.37 & 76.59 & 50.78 \\
\midrule
\multicolumn{12}{l}{\textit{Open-source specialized judge models}} \\
Unified Reward & 55.52 & 64.31 & 50.74 & 56.19 & 5.21 & 52.62 & 49.85 & 83.47 & 87.39 & 77.38 & 48.03 \\
UnifiedReward-Think-qwen-7b & 51.53 & 67.00 & 53.12 & 55.69 & 4.17 & 57.23 & 48.62 & 80.24 & 81.09 & 75.79 & 46.90 \\
R1-Reward & 51.23 & 55.56 & 51.63 & 55.20 & 5.94 & 44.92 & 62.77 & 79.84 & 78.99 & 63.10 & 44.53 \\
\bottomrule
\end{tabular}
}
\end{table*}

\noindent\textbf{Difficulty in discriminating similar-length CoT reasoning pairs.} Models show strong confusion when both positive and negative examples contain comparable CoT lengths and similar reasoning patterns. Experimental results reveal moderate performance across all models on pairwise CoT comparison tasks, with accuracy largely ranging between 50\% and 70\%. Only the top-performing closed-source model, Gemini 2.5 Pro, achieves fairly high performance on this task. Notably, when the chosen–rejected pairs are generated by the same model with similar reasoning length and style, distinguishing between correct and incorrect reasoning becomes significantly more challenging than pairs generated by different models.

\noindent\textbf{Length bias persists under imbalanced pairwise comparisons.} When the lengths of the chosen and rejected responses in a prompt are highly imbalanced, model predictions exhibit significant randomness. For example, in comparisons between CoTs, some models such as GPT-5 and GLM-4.5V display a preference for LongCoT, while others such as Qwen3-VL favor the shorter one. When comparing a short direct answer with an erroneous but logically structured CoT, models consistently prefer the latter, even if it contains reasoning flaws. This indicates a structural and context-dependent length bias that cannot be adequately resolved with existing training paradigms.

\noindent\textbf{Process error judgment remains challenging for small-scale models despite seemingly low task complexity.} Even when confined to a few well-defined categories of process-level errors, current judge models still exhibit notable difficulties in binary classification settings. This challenge is particularly pronounced in the incidental error category, where the high degree of content similarity between positive and negative reasoning chains poses a significant hurdle. It is noteworthy that even specialized judge models, which are explicitly optimized for the judgment task, still demonstrate a considerable performance gap when compared to larger-scale, closed-source counterparts.

\noindent \textbf{Judgment accuracy strongly correlates with base multimodal understanding capabilities.} Large-scale proprietary models generally outperform open-source models. When restricted to small-scale models, newly developed general MLLMs such as Qwen3-VL (4B/8B) outperform several dedicated 7B-scale judge models and achieve SOTA performace on M-JudgeBench, demonstrating that existing judge models primarily overfit on preference-labeled data without acquiring fundamental judging abilities. This limitation holds even for CoT enhanced variants (e.g., UnifiedReward-Think and R1-Reward).

\subsubsection{Performance of M-Judger against baselines}
We evaluate our proposed Judge-MCTS against several strong baselines, including R1-Reward and UnifiedReward-Think, on three benchmarks: (1) M-JudgeBench, (2) VL-RewardBench, and (3) Multimodal RewardBench. Safety tasks are excluded from Multimodal RewardBench because all the test cases selected "Unclear" as the chosen answer, which is not directly related to the judgment capabilities evaluated in this work.

Table \ref{tab:main_results} summarizes the performance of all models on open-source benchmarks and our proposed benchmark (see Supplementary Material for detailed results across all subtasks). Qwen3-VL-8B-Instruct enhanced by Judge-MCTS achieves state-of-the-art performance across all three benchmarks. The results show the superiority of our MCTS-based data construction method. 

\noindent\textbf{Significant improvements in core judge competencies.} Models enhanced by Judge-MCTS (gray rows in Table \ref{tab:main_results}) exhibit large gains in M-JudgeBench. The result indicates that MCTS-augmented reasoning trajectories introduce supervision signals that are directly aligned with human-preferred judgment behavior and grounded in reasoning quality rather than superficial correlations.

\noindent \textbf{No performance degradation on existing judge benchmarks.} Incorporating MCTS-augmented data maintains or improves accuracy on established judge benchmarks, confirming the stability and compatibility of our data generation approach. The strong performance on existing benchmarks confirms that the improvements are general and not specialized to our new benchmark.

\begin{table*}[!htbp]
\centering
\begin{threeparttable}
\caption{Performance comparison of M-Judger series against state-of-the-arts on three judge benchmarks. VL: VL RewardBench; Multimodal: Multimodal RewardBench; Pairwise CoT, Length bias, and Process error: overall accuracy of three main tasks in Table \ref{tab:eval_results_all}, respectively; M-Judger-SFT: Mixing MCTS-augmented data with open-source data during SFT; M-Judger-RL: DAPO training after SFT on only open-source pairwise data.}
\setlength{\tabcolsep}{4.5pt}
\label{tab:main_results}
\begin{tabular}{lcccccc}
\toprule
\multirow{2}{*}{\textbf{Baseline \& Model Variants}} & \multirow{2}{*}{\textbf{VL}} & \multirow{2}{*}{\textbf{Multimodal}} & \multicolumn{4}{c}{\textbf{M-JudgeBench}} \\
\cmidrule{4-7}
& & & \textbf{Pairwise CoT} & \textbf{Length bias} & \textbf{Process error} & \textbf{Overall Acc}\\
\midrule
\multicolumn{5}{l}{\textit{Open-source 7B judge models}} \\
UnifiedReward-Think-qwen-7b & 73.80 & 62.03 & 55.50 & 23.98 & 81.03 & 46.90 \\
R1-Reward & 71.92 & 80.10 & 51.98 & 25.40 & 72.49 & 44.53 \\
\midrule
\multicolumn{5}{l}{\textit{Best models enhanced by Judge-MCTS}} \\
Qwen3-VL-4B-Instruct & 61.75 & 62.98 & 59.02 & 31.49 & 73.71 & 50.00 \\
\quad + SFT (only open-source) & 75.94 & 64.81 & 58.93 & 23.85 & 94.85 & 50.81 \\
\rowcolor{gray!20}
\quad M-Judger-SFT-Qwen4B & 76.50 & 65.43 & 60.40 & 38.14 & 92.14 & 57.00 \\
\rowcolor{gray!20}
\quad M-Judger-RL-Qwen4B & 76.42 & 65.43 & 63.12 & 43.98 & 94.04 & 60.96 \\
Qwen3-VL-8B-Instruct & 63.91 & 65.74 & 63.56 & 24.66 & 84.14 & 50.78 \\
\quad + SFT (only open-source) & 77.15 & 65.41 & 64.59 & 26.15 & 92.55 & 53.48 \\
\rowcolor{gray!20}
\quad M-Judger-SFT-Qwen8B & 76.82 & 65.67 & 66.86 & 34.22 & 90.79 & 57.46 \\
\rowcolor{gray!20}
\quad M-Judger-RL-Qwen8B & 80.75 & 65.52 & 67.45 & 44.53 & 92.14 & 62.42 \\
\bottomrule
\end{tabular}
\end{threeparttable}
\end{table*}

\subsection{Ablation Studies}
\label{sec:ablation}

We perform ablation experiments to understand the contribution of key components on three benchmarks, as shown in Table \ref{tab:main_results}, \ref{tab:ablation} (see Supplementary Material for detailed results across all subtasks). This section aims to examine three central aspects: (1) The limitation of merely expanding SFT data with additional open-source samples. (2) The effectiveness of incorporating MCTS-augmented data. (3) The overall benefit of the two-stage Judge-MCTS framework.

\begin{table*}[!htbp]
\centering
\begin{threeparttable}
\caption{Ablation study of MCTS-augmented SFT data on three judge benchmarks across more base models.}
\setlength{\tabcolsep}{4.5pt}
\label{tab:ablation}
\begin{tabular}{lcccccc}
\toprule
\multirow{2}{*}{\textbf{Baseline \& Model Variants}} & \multirow{2}{*}{\textbf{VL}} & \multirow{2}{*}{\textbf{Multimodal}} & \multicolumn{4}{c}{\textbf{M-JudgeBench}} \\
\cmidrule{4-7}
& & & \textbf{Pairwise CoT} & \textbf{Length bias} & \textbf{Process error} & \textbf{Overall Acc}\\
\midrule
Qwen3-VL-2B-Instruct & 55.25 & 56.17 & 54.99 & 25.96 & 73.04 & 45.99 \\
\quad + SFT (only open-source) & 70.09 & 59.93 & 59.90 & 23.54 & 83.88 & 48.90 \\
\rowcolor{gray!20}
\quad M-Judger-SFT-Qwen2B & 71.13 & 59.36 & 60.19 & 33.79 & 86.18 & 53.91 \\
\midrule
LLaVA-v1.6-Mistral-7B & 39.94 & 57.01 & 47.29 & 40.93 & 52.98 & 45.66 \\
\quad + SFT (only open-source) & 72.57 & 62.34 & 56.38 & 23.23 & 66.80 & 44.07 \\
\rowcolor{gray!20}
\quad M-Judger-SFT-LLaVA7B & 72.33 & 62.38 & 59.53 & 45.96 & 73.71 & 56.47 \\
\midrule
UnifiedReward-qwen-7B & 75.94 & 62.34 & 56.60 & 24.60 & 83.33 & 48.03 \\
\quad + SFT (only open-source) & 81.64 & 63.50 & 62.31 & 24.78 & 88.35 & 51.16 \\
\rowcolor{gray!20}
\quad M-Judger-SFT-Uni7B & 80.91 & 64.03 & 65.32 & 38.70 & 84.82 & 57.60 \\
\midrule
M-Judger-SFT-Intern8B & 51.08 & 63.29 & 56.89 & 22.86 & 70.19 & 47.31 \\
\quad + SFT (only open-source) & 75.62 & 66.12 & 59.90 & 24.97 & 92.55 & 51.24 \\
\rowcolor{gray!20}
\quad M-Judger-SFT-Intern8B & 79.23 & 65.69 & 62.02 & 38.63 & 92.01 & 57.84 \\
\bottomrule
\end{tabular}
\end{threeparttable}
\end{table*}

\noindent\textbf{Effect of open-source pairwise data across different initial backbones.}
Comparing the Qwen3-series models in Table \ref{tab:main_results} with LLaVA-v1.6 and Unified Reward (based on Qwen2.5-VL-7B-Instruct) in Table \ref{tab:ablation} reveals that, with the enhanced capabilities of general-purpose foundation models, the marginal benefit of adding more open-source pairwise data for SFT has largely plateaued, as reflected by their stagnant performance across benchmarks. This suggests that rather than increasing the volume of data by task type, constructing specialized and diverse reasoning data through Judge-MCTS offers a more effective path forward.

\noindent\textbf{Effect of MCTS-augmented data during SFT stage.} 
We inject a proportion of 13k MCTS-augmented pairwise reasoning samples into the 142k open-source mixture. This setup enables a controlled comparison across different initial backbones, isolating how MCTS data improves the SFT stage of judge models. Across all evaluated base backbones, incorporating MCTS pairwise data consistently leads to noticeable performance improvements on M-JudgeBench, particularly in pairwise CoT comparison and length bias avoidance tasks (comparing M-Judger-SFT series with all the five base models in Table \ref{tab:main_results}, \ref{tab:ablation}). These results demonstrate that the structured, multi-step reasoning pairs produced by MCTS provide essential signals for enhancing judgment capabilities. 

\noindent\textbf{Effect of RL stage in Judge-MCTS.} Comparing the M-Judger-RL series with the M-Judger-SFT series in Table \ref{tab:main_results} shows that, although the latter already achieves notable performance gains against "SFT (only open-source)" variants, DAPO further brings more substantial improvements across all evaluation dimensions of M-JudgeBench. The results indicate that RL training is more effective than SFT in learning from limited but high-quality preference data.

\section{Conclusions}
\label{sec:conclusions}

In this work, we introduce a capability-oriented benchmark M-JudgeBench and a data construction framework Judge-MCTS for evaluating and improving multimodal judge models. Unlike prior efforts focused mainly on task categories or final-answer correctness, M-JudgeBench systematically targets both process-level and result-level judgment, reflecting core competencies of reliable human evaluators. It enables deeper analysis of model behavior across reasoning styles, response lengths, and cross-model comparisons. Beyond being a high-quality dataset, M-JudgeBench also provides a generalizable methodology for constructing and upgrading judge benchmarks into more challenging pairwise ranking tasks.

Our evaluation reveals that existing MLLM-as-a-judge systems still exhibit systematic biases: prone to overvalue reasoning fluency, struggle with cross-style adaptation, and remain sensitive to response length. These findings highlight the need to move beyond traditional category-driven evaluation toward ability-driven benchmarking.

To further enhance model capability, we propose Judge-MCTS, which synthesizes structured reasoning trajectories for judge model training. Using Judge-MCTS, we enhanced multiple open-source base models to create the M-Judger series. Empirical results demonstrate consistent gains across existing judge benchmarks as well as M-JudgeBench, validating the effectiveness of Judge-MCTS and the power of M-Judger.

In summary, we construct a unified framework for evaluating and enhancing MLLM-as-a-judge systems. It will serve as a valuable foundation for future research on judge model design and evaluation.


{
    \small
    \bibliographystyle{ieeenat_fullname}
    \bibliography{main}
}

\clearpage
\setcounter{page}{1}
\appendix
\onecolumn
\section*{\centering \Large Supplementary Material}
\tableofcontents
\clearpage
\section{Benchmark Examples}\label{sec:bench_example}
\subsection{Example1. CoTs from the Same Model}
\begin{tcolorbox}[colback=gray!5,colframe=black!50,arc=1mm,boxrule=0.4pt]
\setstretch{1.2}
\textbf{Question: }The following information was taken from the National Accounts of Country Imaginery: From the given figures, the following can be calculated in Rm. What is the GDP at Market Prices?\\
\textbf{Answer: }863,000\\
\textbf{Field: }Economics\\
\textbf{Source: }2023\_Gr10\_Eco\_Oly\\
\textbf{Chosen: }To calculate GDP at Market Prices using the income approach, we sum the following components: \\
1. **Compensation of employees**: 420,000 Rm; 2. **Gross operating surplus**: Net operating surplus + Consumption of fixed capital = 240,000 + 110,000 = 350,000 Rm; 3. **Net taxes on production and imports**: (Taxes on production + Taxes on products) - (Subsidies on production + Subsidies on products) = (16,000 + 84,000) - (2,000 + 5,000) = 100,000 - 7,000 = 93,000 Rm. Adding these together:  [420,000 + 350,000 + 93,000 = 863,000]\\
\textbf{Rejected: }To calculate GDP at Market Prices using the income approach, we sum the following components: \\
1. **Compensation of employees**: 420,000 Rm; 2. **Net operating surplus**: 240,000 Rm; 3. **Consumption of fixed capital** (depreciation): 110,000 Rm; 4. **Net taxes on production** (Taxes on production - Subsidies on production): (16,000 - 2,000 = 14,000) Rm. Adding these together:  [420,000 + 240,000 + 110,000 + 14,000 = 784,000]\\
\textbf{Type: }shortcot\_same\_model\\
\textbf{Explanation: }chosen from glm4.5v, rejected from glm4.5v\\
\includegraphics[width=0.5\linewidth]{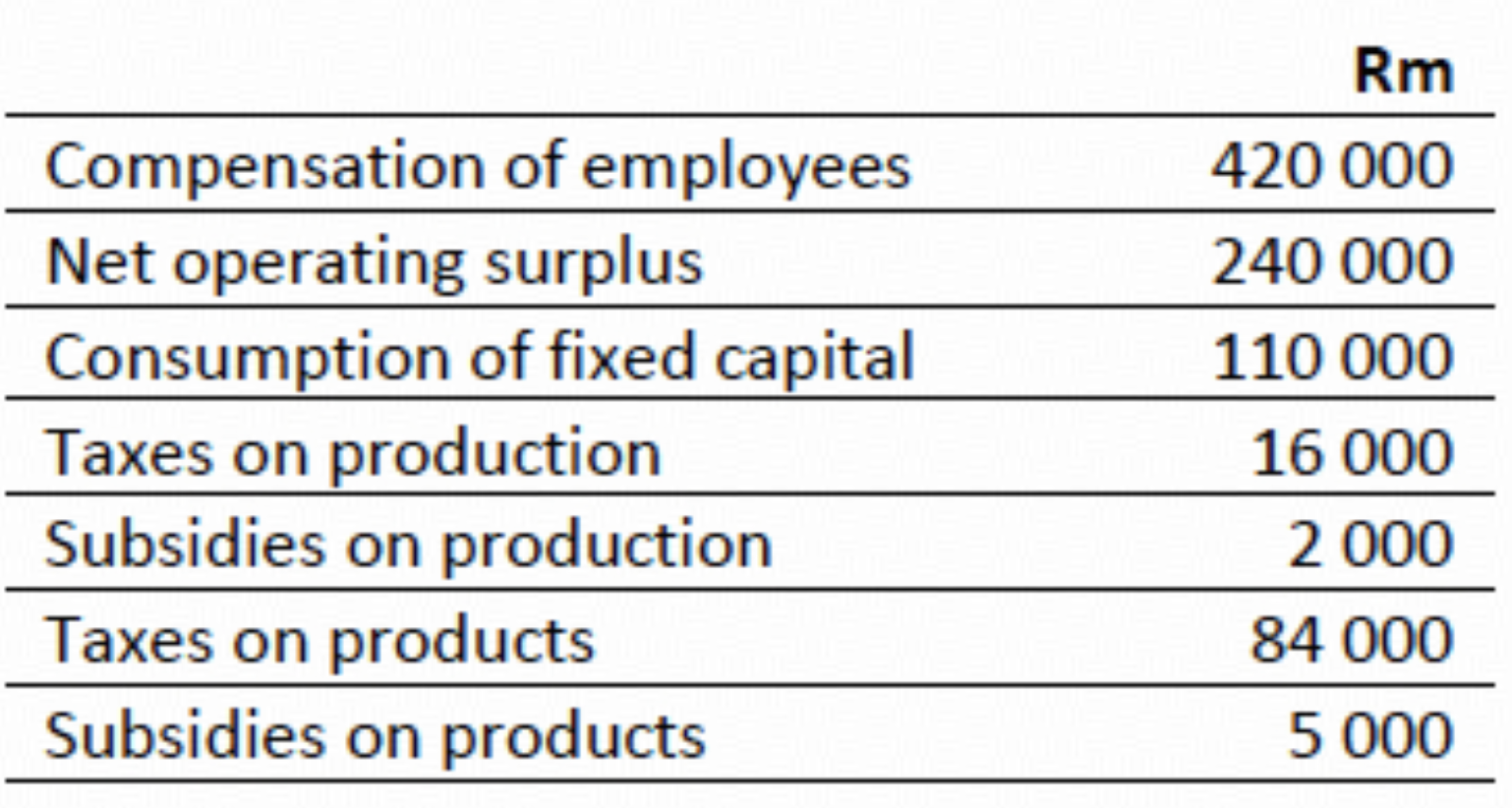}
\end{tcolorbox}
\clearpage

\subsection{Example2. CoTs from Different Models}
\begin{tcolorbox}[colback=gray!5,colframe=black!50,arc=1mm,boxrule=0.4pt]
\setstretch{1.2}
\textbf{Question: }What number should replace the question mark?\\
\textbf{Answer: }6\\
\textbf{Field: }IQ\\
\textbf{Source: }Ultimate-IQ-Test-5-30\\
\textbf{Chosen: }"This is a fun puzzle! Let's solve it by looking for a pattern.\\
1. **Group the numbers by shape.**: **Triangles:** 4, 11, 18; **Circles:** 5, 12, 16; **Squares:** 8, 19, ?\\
2. **Look for a relationship within each group.** A common trick in these puzzles is to check the sum of the numbers for each shape type. **Sum of numbers in triangles: ** 4 + 11 + 18 = 33; **Sum of numbers in circles: ** 5 + 12 + 16 = 33. \\
3. **Apply the pattern.** The pattern is that the sum of the numbers in each shape group is 33. We can apply this rule to the squares. **Sum of numbers in squares: ** 8 + 19 + ? = 33; * 27 + ? = 33; * ? = 33 - 27; * ? = 6\\
The number that should replace the question mark is **6**.\\
\textbf{Rejected: }Let's try to find the pattern based on the shapes and numbers given: Squares: ?, 8, 19. Circles: 12, 5, 16. Triangles: 18, 11, 4. Let's see if the numbers for each shape form a sequence or have something in common. For Squares, known numbers are 8 and 19, in order left-to-right, top-to-bottom: ?, 8, 19. For Circles, known numbers are 12, 5, 16. For Triangles, known numbers are 18, 11, 4. \\
Now, let's check if the numbers for each shape form a sequence. Squares: The difference between 8 and 19 is 11. If we extend backwards: 8 - 11 = -3, which is not reasonable. Circles: The difference between 12 and 5 is -7, and between 5 and 16 is +11. No clear arithmetic sequence. Triangles: 18, 11, 4. 18 - 7 = 11, 11 - 7 = 4. So triangles are decreasing by 7 each time. For circles: 12, 5, 16 → 12 - 7 = 5, 5 + 11 = 16, thus the pattern alternates -7, +11. For squares: if we apply the same alternating pattern, 8 + 11 = 19, and 8 - 7 = 1. So, the sequence would be 1, 8, 19 (if we continue -7, +11). So, the answer is likely 1. \\
Final Answer: Therefore, the number that should replace the question mark is 1. Reasoning: For each shape, the sequence alternates by -7 and +11. Triangles follow -7 consistently, while squares and circles follow -7, +11. Hence, the missing square is 1.\\
\textbf{Type: }shortcot\_diff\_model\\
\textbf{Explanation}: chosen from gemini2.5pro, rejected from gpt4.1\\
\includegraphics[width=0.4\linewidth]{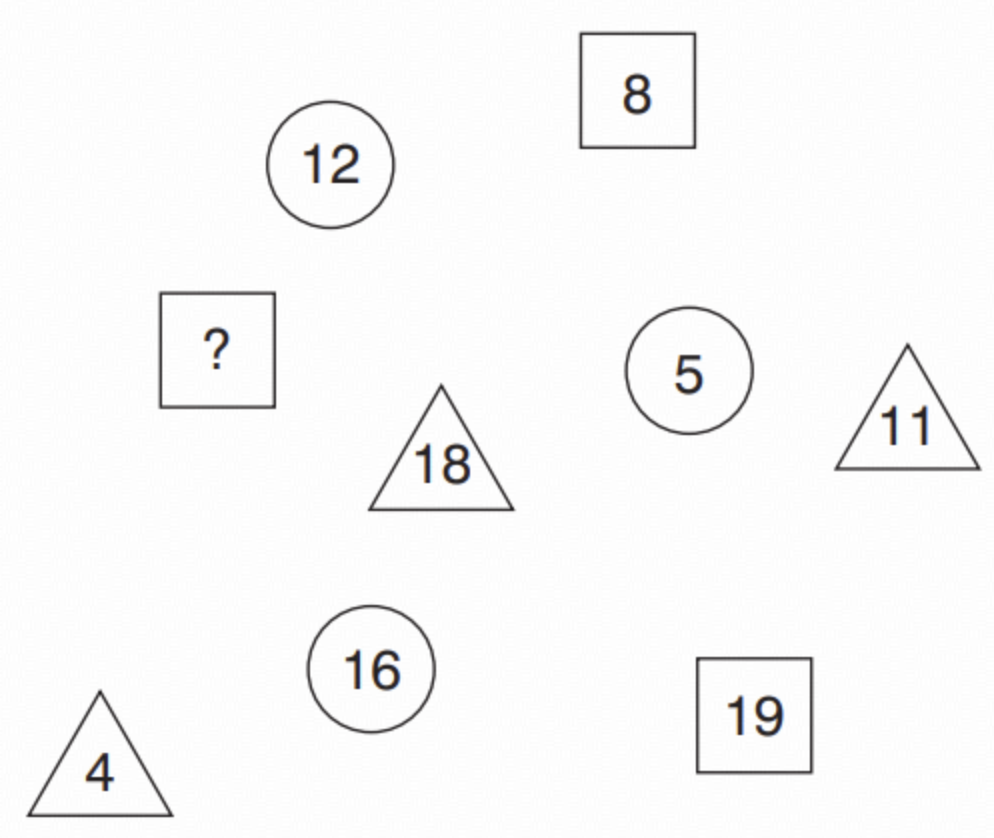}
\end{tcolorbox}
\clearpage

\subsection{Example3. Process Error Judgment}
\noindent\textbf{Example1. Visual Perception Errors}
\begin{tcolorbox}[colback=gray!5,colframe=black!50,arc=1mm,boxrule=0.4pt]
\setstretch{1.2}
\textbf{Question: }The pedigree depicted shows the inheritance of a genotypic characteristic in a family. Circles depict XX genotypes and squares depict XY genotypes. Coloured symbols denote individuals who are afflicted by a genetically inherited disease. What is the most likely type of inheritance of the disease demonstrated by this family tree?\\
\textbf{Answer: }Autosomal recessive\\
\textbf{Field: }Biology\\
\textbf{Source: }Biology Olympiad Past Exams\\
\textbf{Chosen: }To determine the most likely inheritance pattern, we analyze the pedigree as follows: **Key Observations**  1. Unaffected parents (1: XY, 2: XX) have an affected daughter (4: XX) and unaffected children (3: XY, 5: XX); 2. Unaffected parents (5: XX, 6: XY) have an affected son (8: XY) and an unaffected daughter (7: XX).\\
**Analysis of Inheritance Patterns** 1. **Autosomal Dominant**: Unlikely. Affected individuals must inherit the allele from at least one affected parent, but 4 and 8 have unaffected parents. 2. **X-Linked Recessive**: Impossible. For a female (4) to be affected, she must inherit two recessive alleles (one from each parent). However, her father (1) is unaffected, so he can only pass a normal X chromosome, making it impossible for 4 to be affected. 3. **X-Linked Dominant**: Unlikely. Affected individuals must inherit the allele from an affected parent, but 4 and 8 have unaffected parents. 4. **Y-Linked**: Impossible. Only males inherit Y chromosomes, but 4 (female) is affected. 5. **Autosomal Recessive**: Plausible. Unaffected parents (1 and 2) are carriers (Aa). They can have an affected child (4: aa) by passing recessive alleles.  5 (daughter of 1 and 2) is a carrier (Aa, since 25\% chance of being affected, 50\% carrier, 25\% non-carrier). 5 (Aa) and 6 (Aa, a carrier) can have an affected child (8: aa) by passing recessive alleles.\\  
**Conclusion**  The pattern fits **autosomal recessive inheritance**, as unaffected carriers can produce affected offspring of both genders. **Answer:** Autosomal recessive.\\
\textbf{Rejected: }"To determine the most likely inheritance pattern, we analyze the pedigree as follows: **Key Observations**  1. Unaffected parents (1: XY, 2: XX) have an affected **son** (4: XY) and unaffected children (3: XY, 5: XX); 2. Unaffected parents (5: XX, 6: XY) have an affected son (8: XY) and an unaffected daughter (7: XX).  

**Analysis of Inheritance Patterns** ... (\textit{the same as chosen, omitted here.})

**Conclusion**  The pattern fits **autosomal recessive inheritance**, as unaffected carriers can produce affected offspring of both genders. **Answer:** Autosomal recessive."\\
\textbf{Type: }visual\_error\_bygpt\\
\textbf{Explanation}: I introduced a subtle visual interpretation error by changing individual 4’s gender from female (XX) to male (XY) in the “Key Observations” step while keeping their affected status and the rest of the analysis consistent. This small discrepancy alters the perceived visual data from the pedigree without obviously breaking the plausibility of the reasoning, since autosomal recessive inheritance works for either sex. The final conclusion and answer remain unchanged. The modification is minimal and embedded naturally into the comprehension step.\\
\includegraphics[width=0.4\linewidth]{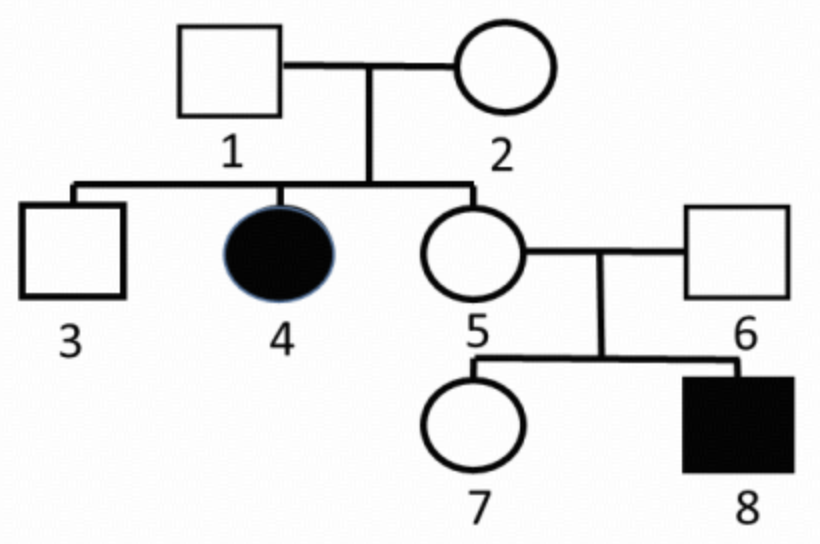}
\end{tcolorbox}
\clearpage
\noindent\textbf{Example2. Logical Reasoning Fallacies}
\begin{tcolorbox}[colback=gray!5,colframe=black!50,arc=1mm,boxrule=0.4pt]
\setstretch{1.2}
\textbf{Question: }What is the name of this organelle?\\
\textbf{Answer: }Mitochondrion\\
\textbf{Field: }Biology\\
\textbf{Source: }Biology Olympiad Past Exams\\
\textbf{Chosen: }The organelle in the image is a **mitochondrion**. Key structural features identifying it include: \\
- A double membrane (outer membrane and folded inner membrane).\\
- The inner membrane forms **cristae** (folds), which are characteristic of mitochondria and increase surface area for energy production.\\
- Its oval/rod-shaped morphology, typical of mitochondria.\\
These features distinguish it from other organelles like chloroplasts (which have thylakoid stacks) or the endoplasmic reticulum (a network of membranes).\\
**Answer:** Mitochondrion."
\\
\textbf{Rejected: }The organelle in the image is a **mitochondrion**. Key structural features identifying it include: \\
- A double membrane, although in this case the outer membrane appears slightly folded inward rather than lying flat, which is sometimes observed in mitochondria under specific conditions. \\
- The inner membrane forms elongated structures that resemble **cristae** (folds), although here they seem parallel and unbranched, which could be confused with certain filament networks but are still typical for mitochondria.\\
- Its elongated oval morphology, which is often associated with mitochondria but can rarely be seen in other organelles under atypical preparation artifacts.\\
These features distinguish it from other organelles like chloroplasts (which have thylakoid stacks and three membranes) or the endoplasmic reticulum (a network-like spread without a defined enclosing membrane pair).\\
**Answer:** Mitochondrion.\\
\textbf{Type: }reasoning\_error\_bygpt\\
\textbf{Explanation}: I introduced a subtle logical flaw by incorrectly describing the outer membrane as “folded inward” and suggesting that cristae appear parallel and unbranched, which could be misinterpreted as other filament structures. This deviates from the usual accurate depiction but still keeps the reasoning close enough to plausible to avoid immediate detection. Additionally, I added the notion that chloroplasts have “three membranes,” which is slightly misleading since they have a double membrane and internal thylakoid membranes, subtly altering the reasoning. The final answer remains unchanged.\\
\includegraphics[width=0.3\linewidth]{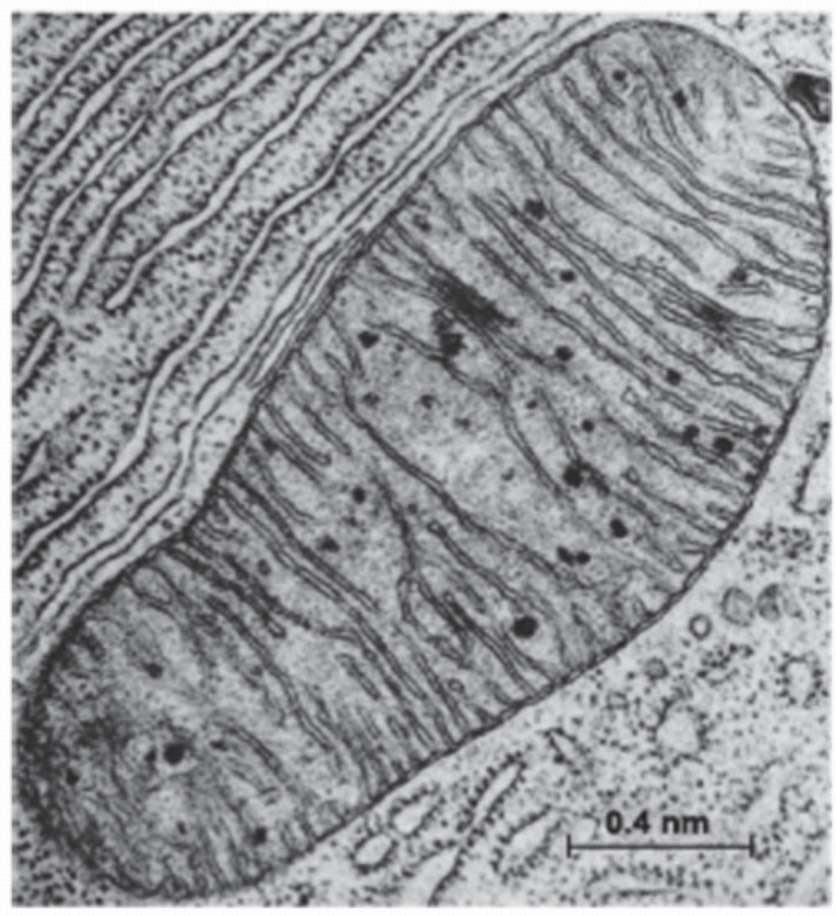}
\end{tcolorbox}
\clearpage

\section{Sources of Benchmark Seed Data}\label{sec:seed_data}
\begin{table}[ht]
\centering
\caption{Statistics of seed data }
\label{tab:seed_data}
\begin{tabular}{lc}
\toprule
\textbf{Source} & \textbf{Number} \\
\midrule
MMMU\cite{yue_mmmu_2024_upd} & 289 \\
MMMU-Pro\cite{yue_mmmu-pro_2025_upd} & 533 \\
MMStar\cite{chen_are_2024_upd} & 96 \\
MMReason\cite{yao_mmreason_2025_upd} & 511 \\
M3CoT\cite{chen_m3cot_2024_upd} & 488 \\
MathVision\cite{wang_measuring_2024_upd} & 548 \\
MathVerse\cite{zhang_mathverse_2024_upd} & 811 \\
\textbf{Total} & \textbf{3276} \\
\bottomrule
\end{tabular}
\end{table}

\section{Sources of Pairwise Training Data}\label{sec:training_data}
\begin{table}[ht]
\centering
\begin{threeparttable}
\caption{Statistics of open-source pairwise training datasets}
\label{tab:opensource_training_data}
\begin{tabular}{lcc}
\toprule
\textbf{Source} & \textbf{Task} & \textbf{Number of Sampled Data\tnote{1}} \\
\midrule
MMPR\cite{wang_enhancing_2025} & Reasoning & 36,679 \\
MMIF\cite{ding_mm-ifengine_2025_upd} & Instruction following & 11,277 \\
RLAIF-V\cite{yu_rlaif-v_2024_upd} & Hallucination & 22,359 \\
POVID\cite{zhou_aligning_2024} & Hallucination & 8,362 \\
MIA-DPO\cite{liu_mia-dpo_2025_upd} & VQA & 13,473 \\
PDS-DPO\cite{wijaya_multimodal_2024} & VQA & 6,160 \\
UnifiedReward-img\cite{wang_unified_2025}\tnote{2} & Mixed & 15,403 \\
Skywork-Reward\cite{liu_skywork-reward_2024_upd} & Mixed & 11,550 \\
StepDPO\cite{lai_step-dpo_2024} & Math & 1,619 \\
Ling-Coder-DPO\cite{codefuse_every_2025} & Coding & 14,914 \\
\midrule
\textbf{Total} & & 141,796 \\
\bottomrule
\end{tabular}
\begin{tablenotes}
{
\footnotesize
\item[1] We perform downsampling on the original data for model training and ablation studies.
\item[2] We integrate the open-source data collected in Unified Reward and select the pairwise data from it, which includes image understanding and image generation judge tasks.
}
\end{tablenotes}
\end{threeparttable}
\end{table}
\clearpage

\section{Prompt Template for Noise Injection}\label{sec:noisy_example}
\subsection{Visual Perception Errors}\label{sec:visual_noise}
\begin{tcolorbox}[colback=gray!5,colframe=black!50,arc=1mm,boxrule=0.4pt]
\setstretch{1.2}
You are a visual comprehension and semantic modification expert. You need to review the instruction and original response, then modify the response to meet these REQUIREMENTS:\\
1. Modification to the original response: Alter the image-related information in the original response to introduce subtle discrepancies from the actual image content, ensure modifications create erroneous visual interpretations while maintaining reasoning plausibility.\\
2. Keep the final answer unchanged: Only modify the visual comprehension process in the original response and make sure the final answer is the same as the original one.\\
3. Principle of minimal modification: Introduce only a small error in the intermediate steps to corrupt the sentence and keep changes minimally detectable to human observers.
\end{tcolorbox}
\subsection{Logical Reasoning Fallacies}\label{sec:logical_noise}
\begin{tcolorbox}[colback=gray!5,colframe=black!50,arc=1mm,boxrule=0.4pt]
\setstretch{1.2}
You are an expert in semantic comprehension and modification. You need to review the instruction and original response, then modify the response to meet these REQUIREMENTS:\\
1. Modification to the original response: Introduce subtle logical flaws or semantic deviations in the reasoning steps while maintaining proximity to correct logic.\\
2. Keep the final answer unchanged: Only modify the reasoning process in the original response and make sure the final answer is the same as the original one.\\
3. Principle of minimal modification: Introduce only a small error in the intermediate steps to corrupt the sentence and keep changes minimally detectable to human observers.
\end{tcolorbox}
\subsection{Incidental Mistakes}\label{sec:incidental_noise}
\begin{tcolorbox}[colback=gray!5,colframe=black!50,arc=1mm,boxrule=0.4pt]
\setstretch{1.2}
You are an expert in semantic comprehension and modification. You need to review the instruction and original response, then modify the response to meet these REQUIREMENTS:\\
1. Error Types to Inject (choose only 1-2 per solution):\\
1.1 Spelling errors: Minor misspellings (e.g., "solution" → "soultion", "calculate" → "calulate")\\
1.2 Numerical errors: Small digit transpositions or value changes (e.g., "12010" → "10210", "3.14" → "3.41")\\
1.3 Content omissions: Skip a short phrase or half-sentence without disrupting overall flow\\
2. Preserve 95 percent of the original text. \\
3. Keep the final answer unchanged: Only modify the process in the original response and make sure the final answer is the same as the original one.\\
4. Make errors appear accidental/natural and ensure the solution remains plausible and logically coherent. \\
Modify the given correct solution text by introducing subtle, hard-to-detect errors. Make minimal changes directly to the original response without rewriting the entire solution.
\end{tcolorbox}
\subsection{Output Format}
For the three types of noise injected in the above subsections (\ref{sec:visual_noise}, \ref{sec:logical_noise}, \ref{sec:incidental_noise}), the model is guided to produce outputs in a unified format to facilitate subsequent extraction.
\begin{tcolorbox}[colback=gray!5,colframe=black!50,arc=1mm,boxrule=0.4pt]
\setstretch{1.2}
For the following conversation between an user and an AI Assistant, output your processed response based on the above rules.\\
\{instruction\}\\
The start of Assistant's Answer\\
\{baseline\_response\}\\
The end of Assistant's Answer

IMPORTANT:
Make sure not to reveal that you are intentionally generating incorrect answers; just output your modifications directly.
Please strictly follow the following format:\\
The start of your processed response\\
\{provide a processed response\}\\
The end of your processed response\\
The start of your explanation to the modification\\
\{explain your modification\}\\
The end of your explanation to the modification
\end{tcolorbox}
\clearpage

\section{Model Evaluation Details}
\subsection{Prompt Template for M-JudgeBench Evaluation}\label{sec:prompt_eval}
\begin{tcolorbox}[colback=gray!5,colframe=black!50,arc=1mm,boxrule=0.4pt]
\setstretch{1.2}
You are given an image and a question related to it. Your task is to evaluate two multimodal reasoning responses based on the following five criteria:\\
1. **Answer Correctness**: Check whether each response provides the most accurate and relevant final answer to the given question. Prioritize factual correctness above all other factors.\\
2. **Reasoning Soundness**: Examine whether the reasoning steps logically lead to the final answer. Identify any reasoning fallacies, invalid inferences, or irrelevant logic chains that might affect reliability.\\
3. **Perceptual Understanding**: Assess the correctness of visual grounding — whether objects, regions, or relationships in the image are correctly interpreted and referenced during reasoning.\\
4. **Conciseness and Coherence**: Evaluate whether the reasoning process is clear, coherent, and appropriately detailed. The response should neither omit essential reasoning nor over-elaborate with redundant or misleading content. Do not prefer longer reasoning by default.\\
5. **Style and Robustness**: Consider whether the model maintains consistent judgment quality across different response styles or model sources. The decision should rely on reasoning quality, not stylistic fluency or writing preference.\\
Please prioritize selecting the response with the most accurate final answer as the chosen one, and then consider the thoroughness and correctness of its reasoning process.\\
After evaluating both responses, clearly state your decision in the format: `Answer 1 is better' or `Answer 2 is better.'\\
Question: \{Query\}\\
Answer 1: \{R1\}\\
Answer 2: \{R2\}\\
Make sure you clearly state your decision only, without any additional explanation.
\end{tcolorbox}
\clearpage

\subsection{Full Evaluation Results on M-JudgeBench}
\label{eval_ablation_allsubtasks}
\begin{table}[!htbp]
\centering
\caption{Performance across all subtasks of M-JudgeBench (supplemental results of Tables in the main text). Short/Long-same/cross: ShortCoT or LongCoT pair from the same model or different models, using correct as chosen and error as rejected; Ans as ch.: correct short answer v.s. incorrect CoT; Short as ch.: correct ShortCoT v.s. incorrect LongCoT; Long as ch.: correct LongCoT v.s. incorrect ShortCoT; Perception: Visual Perception Errors; Reasoning: Logical Reasoning Fallacies; Incidental: Incidental Mistakes.}
\setlength{\tabcolsep}{2.5pt}
\label{tab:eval_ablation_allsubtasks}
\resizebox{\linewidth}{!}{
\begin{tabular}{lccccccccccc}
\toprule
\multirow{2}{*}{\textbf{Model}} & \multicolumn{4}{c}{\textbf{Pairwise CoT comparison}} & \multicolumn{3}{c}{\textbf{Length bias avoidance}} & \multicolumn{3}{c}{\textbf{Process error detection}} & \multirow{2}{*}{\textbf{Overall}} \\
\cmidrule(lr{0.1em}){2-5}\cmidrule(lr{0.1em}){6-8}\cmidrule(lr{0.1em}){9-11}
& \textbf{Short-same} & \textbf{Short-cross} & \textbf{Long-same} & \textbf{Long-cross} & \textbf{Ans as ch.} &  \textbf{Short as ch.} & \textbf{Long as ch.} & \textbf{Perception} & \textbf{Reasoning} & \textbf{Incidental} &  \\
\midrule
Qwen3-VL-4B-Instruct & 58.59 & 71.72 & 55.79 & 59.41 & 3.54 & 76.92 & 40.31 & 89.11 & 92.44 & 73.81 & 50.00 \\
\quad + SFT (only open-source) & 56.75 & 63.97 & 59.05 & 56.44 & 4.69 & 97.85 & 6.46 & 94.76 & 97.90 & 92.06 & 50.81 \\
\rowcolor{gray!20}
\quad M-Judger-SFT-Qwen4B & 62.88 & 65.99 & 59.94 & 54.21 & 25.83 & 96.62 & 16.00 & 94.35 & 97.90 & 84.52 & 57.00 \\
\rowcolor{gray!20}
\quad M-Judger-RL-Qwen4B &  69.02 & 64.98 & 61.72 & 58.17 & 32.40 & 80.31 & 41.85 & 95.16 & 97.48 & 89.68 & 60.96 \\
\midrule
Qwen3-VL-8B-Instruct & 57.66 & 80.81 & 58.75 & 61.88 & 2.71 & 65.23 & 52.00 & 90.32 & 95.37 & 76.59 & 50.78 \\
\quad + SFT (only open-source) & 57.06 & 73.40 & 63.20 & 65.35 & 4.27 & 90.15 & 26.77 & 95.16 & 97.06 & 85.71 & 53.48 \\
\rowcolor{gray!20}
\quad M-Judger-SFT-Qwen8B & 64.72 & 75.42 & 63.80 & 64.85 & 15.31 & 83.69 & 40.62 & 93.55 & 97.06 & 82.14 & 57.46 \\
\rowcolor{gray!20}
\quad M-Judger-RL-Qwen8B & 66.26 & 76.43 & 66.17 & 63.86 & 35.63 & 73.23 & 51.38 & 91.53 & 95.38 & 83.73 & 62.93 \\
\midrule
Qwen3-VL-2B-Instruct & 46.93 & 59.93 & 60.53 & 53.22 & 8.02 & 59.08 & 45.85 & 79.44 & 87.82 & 52.78 & 45.99 \\
\quad + SFT (only open-source) & 55.52 & 70.71 & 61.72 & 53.96 & 1.88 & 88.92 & 22.15 & 79.44 & 89.08 & 83.33 & 48.90 \\
\rowcolor{gray!20}
\quad M-Judger-SFT-Qwen2B & 57.36 & 69.02 & 59.35 & 56.68 & 17.50 & 87.38 & 28.31 & 86.69 & 94.96 & 77.38 & 53.91 \\
\midrule
LLaVA-v1.6-Mistral-7B & 46.31 & 47.81 & 48.51 & 46.90 & 35.52 & 52.47 & 45.54 & 53.63 & 57.14 & 48.41 & 45.66 \\
\quad + SFT (only open-source) & 56.13 & 64.98 & 54.76 & 51.86 & 3.33 & 87.96 & 17.54 & 61.69 & 72.69 & 66.27 & 44.07 \\
\rowcolor{gray!20}
\quad M-Judger-SFT-LLaVA7B & 57.98 & 69.36 & 58.04 & 55.09 & 41.04 & 89.51 & 17.23 & 70.56 & 89.50 & 61.90 & 56.47 \\
\midrule
UnifiedReward-qwen-7B & 55.52 & 64.31 & 50.74 & 56.19 & 5.21 & 52.62 & 49.85 & 83.47 & 87.39 & 77.38 & 48.03 \\
\quad + SFT (only open-source) & 58.90 & 68.01 & 63.20 & 59.65 & 2.92 & 88.62 & 25.54 & 88.71 & 94.54 & 82.14 & 51.20 \\
\rowcolor{gray!20}
\quad M-Judger-SFT-Uni7B & 64.11 & 71.38 & 62.91 & 63.37 & 23.65 & 79.08 & 42.77 & 84.68 & 92.44 & 77.78 & 57.60 \\
\midrule
InternVL3.5-8B & 55.83 & 60.27 & 57.86 & 52.48 & 5.73 & 52.62 & 58.77 & 85.48 & 87.82 & 59.52 & 47.31 \\
\quad + SFT (only open-source) & 61.66 & 66.33 & 57.27 & 55.94 & 3.02 & 75.69 & 39.08 & 89.11 & 95.80 & 92.86 & 51.24 \\
\rowcolor{gray!20}
\quad M-Judger-SFT-Intern8B & 64.72 & 70.03 & 57.86 & 57.43 & 23.85 & 79.38 & 41.53 & 90.32 & 97.06 & 88.89 & 57.84 \\
\bottomrule
\end{tabular}
}
\end{table}

\clearpage

\section{Reproducible Experimental Details}
\subsection{M-JudgeBench Evaluation}
The benchmark data and evaluation scripts are included in \href{https://github.com/czythu/M_Judger}{https://github.com/czythu/M\_Judger}. Model inference primarily relies on vLLM\cite{kwon2023efficient} and transformers\cite{wolf-etal-2020-transformers} (vllm 0.11.0 and transformers 4.57.0 for Qwen3-VL series).
\subsection{M-Judger Training}
The SFT stage is implemented with LLaMA-Factory\cite{zheng2024llamafactory} and the RL stage is conducted using EasyR1\cite{zheng2025easyr1}. The training details will be made publicly available soon, including:
\begin{itemize}
    \item Training data: 142k open-source collection and 13k MCTS-augmented data.
    \item Fine-tuned models for main results and ablation studies.
\end{itemize}

\end{document}